\pdfoutput=1
\documentclass[11pt]{article}

\usepackage[final]{acl}

\usepackage{times}
\usepackage{latexsym}
\usepackage{booktabs}
\usepackage{multirow}
\usepackage{amsmath}
\usepackage{float}

\usepackage[T1]{fontenc}
\usepackage[utf8]{inputenc}

\usepackage{microtype}

\usepackage{inconsolata}

\usepackage{graphicx}

\title{Mitigating Object Hallucination via Robust Local Perception Search}

\author{
  \textbf{Zixian Gao\textsuperscript{1,2}}, 
  \textbf{Chao Yang\textsuperscript{1$\dagger$}}, 
  \textbf{Zhanhui Zhou\textsuperscript{1}}, 
  \textbf{Xing Xu\textsuperscript{2}}, 
  \textbf{Chaochao Lu\textsuperscript{1}} \\
  \textsuperscript{1}Shanghai Artificial Intelligence Laboratory \\
  \textsuperscript{2}Center for Future Media \& School of Computer Science and Engineering, \\University of Electronic Science and Technology of China  \\
  \texttt{\{zixian.gaoo, asap.zzhou\}@gmail.com},\\ \texttt{\{yangchao, luchaochao\}@pjlab.org.cn}, \texttt{xing.xu@uestc.edu.cn}
}
\begin{document}
\maketitle

\def\thefootnote{$\dagger$}\footnotetext{Corresponding author. Work done while ZG and ZZ were at Shanghai AI Lab.}

\begin{abstract}
Recent advancements in Multimodal Large Language Models (MLLMs) have enabled them to effectively integrate vision and language, addressing a variety of downstream tasks. 
However, despite their significant success, these models still exhibit hallucination phenomena, where the outputs appear plausible but do not align with the content of the images. 
To mitigate this issue, we introduce Local Perception Search (LPS), a decoding method during inference that is both simple and training-free, yet effectively suppresses hallucinations. 
This method leverages local visual prior information as a value function to correct the decoding process. 
Additionally, we observe that the impact of the local visual prior on model performance is more pronounced in scenarios with high levels of image noise. Notably, LPS is a plug-and-play approach that is compatible with various models. 
Extensive experiments on widely used hallucination benchmarks and noisy data demonstrate that LPS significantly reduces the incidence of hallucinations compared to the baseline, showing exceptional performance, particularly in noisy settings. Code is available at \url{https://github.com/ZixianGao/Local-Perception-Search}.
\end{abstract}

\section{Introduction}
\label{sec:intro}

In recent years, Multimodal Large Language Models (MLLMs) have experienced rapid development \cite{du2022survey,ghosh2024exploring,gupta2023cliptrans,zhu2023minigpt}, profoundly transforming the field of multimodal learning. These models have demonstrated remarkable capabilities across a wide range of tasks, including image captioning \cite{zhang2024ferret,kim2023mitigating}, visual question answering \cite{antol2015vqa,kamalloo2023evaluating}, multimodal reasoning \cite{ma2023crepe,qi2024quantifying}, etc. However, despite such advancements, MLLMs are not yet entirely reliable. In real-world scenarios—especially when the visual modality is subjected to adversarial perturbations—MLLMs often suffer from object hallucination \cite{zhang2024multitrust,gao2024embracing}, where the model erroneously generates non-existent objects in its outputs, resulting in misalignment between the generated text and the visual content.

Object hallucination remains a persistent challenge that limits the deployment of MLLMs in safety-critical applications. It has garnered increasing attention in recent years. Expanding the scale and quality of training data \cite{zhao2024lova3,fu2024mme} has proven to be an effective approach for enhancing model performance and reducing hallucinations. However, this strategy also incurs significant annotation costs and leads to considerable computational overhead during training, thereby hampering scalability. In response, recent studies have explored inference-time search techniques as a promising complementary strategy to improve response quality without retraining.
\begin{figure}[t]
  \centering
  \includegraphics[width=1.0\linewidth]{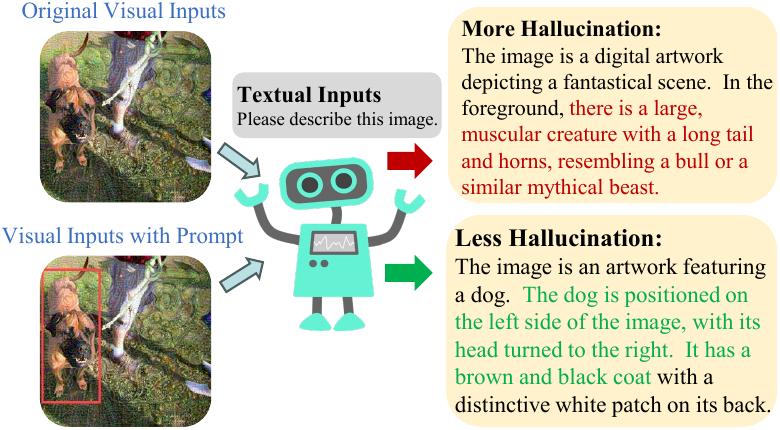}
  \caption{We observed a consistent phenomenon across various MLLMs: when fed with the original visual input alone, the models exhibited hallucinations caused by the adversarial perturbations in the image. In contrast, when provided with a visual input accompanied by a prompt (e.g., a bounding box), the models were able to correctly recognize the object in the image as a dog.}
  \label{fig:intro}
\end{figure}

Several works \cite{zhou2024weak,liu2024inference,kim2024code} have proposed inference-time search algorithms aimed at mitigating hallucinations. For example, \cite{wang2024scaling} formulates the search process as a Markov Decision Process and employ a high-quality Process Reward Model (PRM) trained on curated data to guide decoding. \cite{chen2024halc} adopts contrastive decoding strategies that leverage pre-trained grounding detectors to identify objects in the image and provide auxiliary visual grounding. While these methods show potential, each has its limitations: the former requires training a dedicated PRM, which involves costly annotation and computational demands; the latter depends heavily on external grounding detectors, reducing scalability and limiting real-world applicability. These drawbacks motivate our first research question: \emph{Can we design an inference-time search strategy that relies solely on the internal capabilities of the MLLM, without external supervision or auxiliary models?}

Moreover, recent research \cite{pi2024mllm,guo2024vllm,ding2024eta,zhang2024seeing} has highlighted the vulnerability of the visual modality in MLLMs. Much of the model’s capacity is inherited from the LLM backbone, and insufficient vision-language alignment renders the visual modality particularly fragile to adversarial attacks. Although the aforementioned techniques alleviate hallucinations to some extent, they remain ineffective when the visual input is compromised. As shown in Fig. \ref{fig:intro}, an untargeted adversarial attack leads to severe image distortion, yet the model continues to hallucinate non-existent objects (e.g., hallucinated animals in red text), failing to recognize the actual species in the image. Recent efforts such as \cite{fang2024uncertainty} have proposed uncertainty-aware decoding strategies, quantifying visual uncertainty and using voting-based ensembles to enhance response reliability. However, these methods sacrifice performance on clean images and are computationally expensive due to repeated inference. This leads us to our second research question: \emph{Is it possible to perform trustworthy inference-time intervention that preserves model performance on benign inputs while mitigating hallucinations under adversarial attacks?}

To address these two challenges, we propose a novel locally-aware prior-guided search strategy. Our method exploits the MLLM's intrinsic ability to focus on local visual regions in order to generate priors that guide decoding, resulting in robust and high-quality responses. Specifically, as illustrated in Fig. \ref{fig:intro}, we observe that when bounding boxes are overlaid on adversarially perturbed images, the model successfully identifies the object (e.g., a dog). However, requiring manual or model-generated bounding boxes contradicts our first research objective. Motivated by this observation, we further explore the model's local visual attention capabilities and find that the MLLM can correctly identify objects by attending to localized regions, even without explicit bounding box supervision. We leverage this insight to construct priors from the model’s own attention over local patches, enabling inference-time guidance without reliance on external tools or annotations. This approach allows for efficient and effective hallucination mitigation while maintaining and preserving performance across diverse visual conditions.

Our contributions can be summarized as follows:
\begin{itemize}
    \item We identify and leverage the intrinsic local perceptual capability of MLLMs, demonstrating that this property can be exploited to robustly guide the decoding process even when the visual modality is subjected to adversarial perturbations.
    \item We propose LPS (Local Perception Search), a plug-and-play inference-time search algorithm that utilizes the model’s local visual perception without requiring external supervision. LPS significantly reduces object hallucination across various scenarios.
    \item We conduct comprehensive evaluations on widely adopted hallucination benchmarks, verifying the effectiveness of our approach. Furthermore, we collect and extend an adversarially perturbed hallucination dataset to validate the robustness of LPS under attack conditions.
\end{itemize}

\section{Related Work}

\textbf{Multimodal Large Language Models.}
% https://arxiv.org/pdf/2501.01926
In recent years, the rapid advancement of Large Language Models (LLMs) \cite{touvron2023llama,yang2024qwen2} has also driven significant innovations in the field of multimodal learning. Multimodal Large Language Models (MLLMs) \cite{hurst2024gpt,li2024llava,abdin2024phi}, which employ LLMs as the backbone, have emerged as the dominant paradigm for addressing vision-language tasks. Early works such as LLaVA \cite{liu2023visual} and InstructBLIP \cite{dai2023instructblipgeneralpurposevisionlanguagemodels} adopted visual instruction tuning and achieved effective alignment between visual and linguistic semantic spaces, thereby enabling unified decoding across both modalities. With the growing volume and quality of training data, alongside continuous algorithmic refinements, MLLMs have demonstrated remarkable capabilities. However, despite their impressive performance, these models are not always reliable—particularly when the visual modality is subjected to adversarial perturbations \cite{zhang2024multitrust}. Under such conditions, MLLMs often exhibit hallucinations \cite{chen2024halc,kim2024code}, generating textual outputs that are inconsistent with the visual content, which significantly constrains their deployment in safety-critical applications. Our work aims to mitigate the hallucination issues prevalent in current MLLMs and to facilitate their reliable deployment across a wide range of application domains.

\noindent
\textbf{Hallucination in MLLMs.}
Object hallucination (OH) has remained a persistent challenge in Multimodal Large Language Models (MLLMs), drawing increasing attention in recent years. Prior studies \cite{gunjal2024detecting,zhai2023halle} categorize OH into three primary types: object existence hallucination, attribute hallucination, and relationship hallucination. Recently, a number of works \cite{li2023evaluatingobjecthallucinationlarge,rohrbach2019objecthallucinationimagecaptioning,wang2023amber} have focused on evaluating and detecting OH. For example, POPE \cite{li2023evaluatingobjecthallucinationlarge} formulates OH as a binary classification problem, asking the model to determine whether a specific object exists in the image. In parallel, substantial efforts have been made to develop methods aimed at reducing hallucinations in MLLMs, including both training-based \cite{gunjal2024detecting} and inference-time \cite{wang2024scaling} approaches. However, these methods often suffer from significant resource overhead or rely heavily on prior knowledge provided by external models, which limits their scalability and applicability in real-world settings. This consistent improvement clearly demonstrates that LPS is effective not only across different model architectures but also remarkably robust to changes in model capacity, further highlighting its scalability and broad general real-world applicability.

\noindent
\textbf{Inference-time search.}
% https://arxiv.org/pdf/2412.03704
Inference-time search \cite{zhou2024weak,li2024common,tian2024toward} has been widely adopted in the field of Large Language Models (LLMs), playing a pivotal role in reasoning and hallucination reduction. Given the widespread use of LLM backbones in Multimodal Large Language Models (MLLMs), these search strategies have also gained considerable traction in multimodal contexts \cite{kim2024code, chen2024halc}. A key component of such strategies is the provision of a high-quality reward signal during search. Existing methods typically rely either on external models—such as CLIP \cite{zhou2024calibrated}—to provide reward signals, or on training a dedicated Process Reward Model (PRM) \cite{wang2024scaling} using curated datasets to score candidate outputs. However, these approaches often offer limited effectiveness or incur significant computational overhead, and their applicability in MLLMs remains underexplored. In our work, we leverage the intrinsic capabilities of the model itself to generate prior information and derive reward signals, thereby enhancing the robustness and generalization of MLLMs across diverse scenarios.

\section{Method}

\subsection{Preliminaries}

\subsubsection{Problem Formulation}
We first introduce the phenomenon of object hallucination. Consider an MLLM $M_{\theta}$ with parameters $\theta$. 
This model takes a prompt-image pair $(x,I)$ as input and autoregressively decodes it into the corresponding text $y$. 
Formally, we can express this mathematically as:
\begin{equation}
    y_{t} \sim p_{\theta}(\cdot|v,x,y_{<t}) \propto exp f_{\theta}(\cdot|v,x,y_{<t})
\end{equation}
Where $y$ represents the token at the $t-th$ step, and $y_{<t}$ denotes the sequence of tokens generated up to time step $t$. 
$f$ is the logit distribution generated by the $M_{\theta}$.
Object hallucination occurs when certain parts of the generated text $y$ do not align with the input image $I$, and this phenomenon becomes more pronounced, especially when $I$ is attacked. Our ultimate goal is to mitigate the hallucination phenomenon while effectively preventing image attacks from exacerbating it, all while consistently maintaining high-quality text generation.

\begin{figure*}[t]
  \centering
  \includegraphics[width=1.0\linewidth]{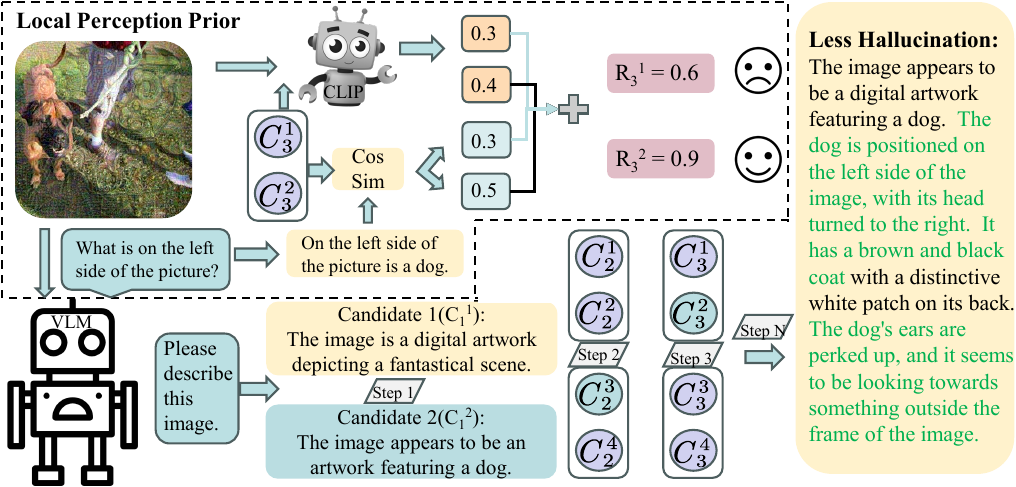}
  \caption{Overall inference framework of our method. The dashed box highlights the Local Perception Prior process, which refines candidate selection based on local visual cues. The remaining components depict the search procedure during a single inference pass. Candidates highlighted with a green background indicate the selected choice at each time step.}
  \label{fig:framework}
\end{figure*}

\subsubsection{Formulation of MLLM Inference}
\label{inference}
We input a prompt-image pair and generate the corresponding $y= [y_{1},y_{2},...,y_{m}]$, where $y$ consists of $m$ step-level responses. Each step-level response $y$ is considered a sample drawn from the conditional probability distribution $y_{t}= p_{\theta}(\cdot|v,x,y_{<t})$. In this paper, we use sentence-level responses, where each step outputs a sentence. Therefore, the text generation task can be formulated as a Markov decision process problem. In this process, a reward function $R$ evaluates the reward for each action, which is also referred to as the process reward model (PRM) in large language models (LLMs). A better reward function tends to yield better responses $y$. Many methods have been proposed in this area, such as guiding large models with smaller models or training a better value model. Our approach aims to leverage the inherent capabilities of MLLMs to guide the search process, thereby improving the quality of generation and saving computational resources.

\subsection{Local Perception Abilities in MLLMs}

As demonstrated in \ref{sec:intro}, we have found that visual prompts play a significant role in assisting Multimodal Large Language Models (MLLMs) in recognizing objects within images, especially in noisy conditions. 
However, applying such visual prompts in practice is challenging. One approach involves manual annotation, while another relies on specialized models for auxiliary labeling. 
The former requires substantial human labor, making it impractical for real-world applications. 
The latter not only incurs additional computational costs but also fails to accurately label visual prompts in noisy conditions. 
Therefore, we are exploring more efficient and precise methods to integrate visual priors.

Through further exploration, we discovered that MLLMs inherently possess a capability similar to the assistance provided by visual prompts in perceiving objects within images. 
We refer to this capability as local perception ability.
As shown in Fig. \ref{fig:framework}, we applied an untargeted attack \cite{chen2023rethinking} to the image, resulting in blurring and distortion. In this case, MLLMs struggle to discern the features of objects within the image. 
When prompted with "Please list the objects in this image," the model is unable to correctly identify or describe the objects present. 
However, when prompted with "Please list the object on the left of this image," the model successfully identifies that there is a dog on the left side of the image. 
This observation suggests that the model inherently possesses the ability to focus on local objects. Moreover, this ability enhances the model's reliability, helping it to mitigate the effects of such attacked noise and enabling accurate object recognition. 
This finding lays the foundation for developing safe and trustworthy inference models.

\subsection{Local Perception Prior Acquisition}
From the autoregressive inference process of MLLMs, it is clear that a highly capable reward function is essential for providing correct feedback, which helps in object recognition within images and facilitates precise decoding. However, existing methods \cite{chen2024halc,wang2024scaling} either require training a reward model from scratch to provide feedback or depend on an auxiliary model to offer prior information for comparative decoding. Both approaches incur additional resource consumption. Therefore, we seek an efficient and resource-free method for constructing the reward function. As mentioned in the previous section, MLLMs inherently possess local perceptual capabilities. Inspired by this, we exploit this capability of MLLMs to generate prior information through the model itself.

As previously mentioned, when the visual modality input to Multi-modal Large Language Models (MLLMs) is subjected to adversarial attacks, the models often fail to recognize the objects present in the image, resulting in hallucinations. However, as illustrated in Figure X, when the model's visual attention is constrained to localized regions, it is still able to correctly identify the objects even under attack. Leveraging this property, we generate prior information to assist the model. Specifically, we employ the following prompt:
\begin{quote}
    x= "Please carefully observe the top, bottom, left, and right parts of the image in sequence, and list the objects present in each section."
\end{quote}

Next, we feed the target image along with the prompt into the MLLM, allowing the model to first generate prior information by leveraging its global-local perception capability before proceeding with the primary task:
\begin{equation}
    y = M_{\theta}(x,I),
\end{equation}
where $y$ denotes the textual prior information generated by the model.
% $$x= "Please carefully observe the top, bottom, left, and right parts of the image in sequence, and list the objects present in each section."$$

\subsection{Calibrated Decoding Search Process}
In \ref{inference}, we define the inference process of Multimodal Large Language Models (MLLMs). We formulate the text generation task as a Markov Decision Process (MDP), characterized by a tuple $(S,A,R,\gamma)$. Here, $S$ denotes the state space, where each state represents a combination of the generated sentence and the associated image. The initial state $s_{0}$ corresponds to the input image $I$ and textual prompt $x$. A represents the action space, in which each action corresponds to the sentence generated at a given step. The reward function $R$ evaluates the quality of each generated action, and $\gamma$ denotes the discount factor.

In this work, we aim to design a reward function $R$ that is both resource-efficient and robust. For the conditional distribution $p_{\theta}(\cdot|x,I,y_{<i},T_{n})$, we sample $k$ candidate outputs at each generation step $t$, denoted as $\hat{y_{t}}=[c^{1}_{t},c^{2}_{t},...,c^{k}_{t}]$. Each candidate $c^{i}_{t}$ is evaluated using the object prior information $y$, which is extracted in the previous stage. Specifically, we compute the cosine similarity between the candidate embedding and the prior embedding as the reward: 
\begin{align}
R^{t}_{LPS}(i) &= \text{CosSim}(y, c_t^{i}) \notag \\
         &= \frac{y \cdot c_t^{i}}{\|y\| \cdot \|c_t^{i}\|}, \quad i = 1, 2, \dots, k.
\end{align}
We define $R^{t}_{LPS}(i)$ as the reward assigned to the $i-th$ candidate at generation step $t$, representing the degree of semantic alignment between the candidate output and the object-level prior knowledge. This similarity score serves as the per-step reward, providing a semantic evaluation signal that guides the model toward generating outputs that are more consistent with the extracted object-level priors.

In addition, to ensure that each generated candidate is semantically relevant to the input image, we incorporate a second alignment signal based on vision-language similarity. Specifically, we compute the cosine similarity between each candidate sentence and the image using CLIP embeddings. Formally, for each candidate $c^{i}_{t}$, we define the CLIP-based image-text similarity score as:
\begin{align}
R_{\text{CLIP}}^t(i) = \text{CLIP}(I,c^{i}_{t}).
\end{align}

This dual-alignment strategy—leveraging both object-level prior matching and global image-text matching—enables us to generate text that is not only semantically aligned with extracted priors but also consistent with the visual content of the image. To integrate these two signals, we define the final reward function as a weighted combination of the prior-based and CLIP-based similarities:
\begin{equation}
    R^i_t = \alpha R_{\text{LPS}}^t(i) + \beta R_{\text{CLIP}}^t(i)
    \label{equal:hyper}
\end{equation}
where $\alpha$ and $\beta$ control the relative contribution of local prior matching and global vision-language alignment. Such a design helps mitigate hallucination phenomena commonly observed in multimodal generation tasks by encouraging both local semantic consistency and global visual alignment.

At each generation step $t$, the candidate $c^{i}_{t}$ with the highest reward $R^i_t$ is selected, ensuring minimal hallucination and strong semantic alignment with the visual input. This step-wise selection process is repeated throughout the sequence generation, ultimately yielding the final output $\hat{y}$.

\begin{table*}[t]
  \centering
  \resizebox{1.0\textwidth}{!}{
    \begin{tabular}{l|ll|ll|ll|ll|llll}
    \toprule
    \multirow{3}{*}{Method} 
      & \multicolumn{8}{c|}{POPE} 
      & \multicolumn{4}{c}{CHAIR} \\
    \cline{2-13}
     ~  & \multicolumn{2}{c|}{adversarial} 
         & \multicolumn{2}{c|}{popular} 
         & \multicolumn{2}{c|}{random} 
         & \multicolumn{2}{c|}{overall} 
         & \multirow{2}{*}{$\mathrm{C}_{\mathrm{s}}$ $\downarrow$}  
         & \multirow{2}{*}{$\mathrm{C}_{\mathrm{i}}$ $\downarrow$}  
         & \multirow{2}{*}{$\mathrm{B}_{\mathrm{1}}$ $\uparrow$}  
         & \multirow{2}{*}{$\mathrm{B}_{\mathrm{4}}$ $\uparrow$}  \\
    \cline{2-9}
     ~  & Acc $\uparrow$ & F1 $\uparrow$ & Acc $\uparrow$ & F1 $\uparrow$ & Acc $\uparrow$ & F1 $\uparrow$ & Acc $\uparrow$ & F1 $\uparrow$ 
         &  &  &  &  \\
    \hline
    Qwen 2.5 VL+  CLIP PRM  & 82.2 & 78.8  & 82.9 & 79.8 & 83.3 & 80.1 & 82.8 & 79.6 & 24.0 & 8.6 & 4.4 & 0.1 \\
    \textbf{Qwen 2.5 VL+ LPS}  & \textbf{83.7} & \textbf{81.2} & \textbf{84.4} & \textbf{81.9} & \textbf{85.4} & \textbf{83.0} & \textbf{84.5} & \textbf{82.0} & \textbf{20.4}  & \textbf{7.9} & \textbf{8.1} &  \textbf{1.7} \\
    \hline
    Llama 3.2 Vision + CLIP PRM & 82.4 & 83.8 & 85.8 & 86.5  & 89.8 & 89.8 & 86.0 & 86.7 & 24.0 & 10.3 & 3.0 & \textbf{0.7}\\
    \textbf{Llama 3.2 Vision + LPS}  & \textbf{83.0} & \textbf{84.3} & \textbf{86.0} & \textbf{86.7} & \textbf{90.7} & \textbf{90.7} & \textbf{86.6} & \textbf{87.2} & \textbf{22.2}  & \textbf{8.9} & \textbf{3.1} &  \textbf{0.7} \\
    \hline
    Phi 3.5 Vision + CLIP PRM & 81.4 & 81.8 & 81.7 & 81.5  & 84.8 & 84.8 & 82.7 & 82.7 & 22.2 & 10.1 & \textbf{14.7} & 2.9 \\
    \textbf{Phi 3.5 Vision + LPS}  & \textbf{82.7} & \textbf{82.1} & \textbf{83.8} & \textbf{82.8} & \textbf{85.8} & \textbf{84.6} & \textbf{84.1} & \textbf{83.2} & \textbf{21.2}  & \textbf{8.5} & \textbf{14.7} &  \textbf{3.1} \\
    \bottomrule
    \end{tabular}
  }
  \caption{\label{tab:pope_chair}Results on POPE and CHAIR. CLIP-PRM denotes the baseline using CLIP as the reward model, while LPS is our proposed method. Best performance in each setting is shown in bold.}
\end{table*}

\section{Experiments}
\subsection{Experimental Settings}
\textbf{Experimental Setup. } 
To validate the effectiveness of our proposed method across different Multimodal Large Language Models (MLLMs), we implemented our approach on the following models: Llava 1.5 (7B, 13B) \cite{liu2024llava1_5}, Qwen 2.5 VL(7B) \cite{yang2024qwen2}, LLaMA 3.2-V (11B) \cite{grattafiori2024llama}, and Phi 3.5-V (4B) \cite{abdin2024phi}. We conducted comparative evaluations against conventional baselines that employ CLIP as the process reward model. In addition, we conducted further comparative experiments on selected datasets using the Visual Contrastive Decoding (VCD) \cite{leng2023mitigatingobjecthallucinationslarge} strategy.

\noindent
\textbf{Benchmarks and Evaluation Metrics.}

Following standard protocols \cite{leng2023mitigatingobjecthallucinationslarge,chen2024halc}, we used two benchmark datasets—POPE \cite{li2023evaluatingobjecthallucinationlarge} and CHAIR \cite{rohrbach2019objecthallucinationimagecaptioning}—to evaluate model robustness against hallucination. POPE targets object-level hallucination in VQA, comprising binary questions like “Is there a dog in the image?”, formed via image-object annotation triplets. CHAIR focuses on caption-level hallucination by measuring the presence of non-existent objects in generated captions. It includes CHAIR-S (sentence-level hallucination rate) and CHAIR-I (instance-level rate). For CHAIR, we sampled 500 images from the MSCOCO 2017 validation set \cite{lin2014microsoft}.

Furthermore, to evaluate model robustness under noisy conditions, we followed the setup in Multitrust \cite{zhang2024multitrust} and extended their untargeted attack dataset. Specifically, we tested models and methods under adversarially perturbed visual modality. The attacks used in this setting are black-box in nature and follow the SSA-CWA \cite{chen2023rethinking} strategy proposed in [24], which is known for its high transferability to unseen MLLMs.
Let $\left\{ f_i^v \right\}_{i=1}^{N}$ represent the vision encoders of a set of surrogate models. The adversarial attack objective is formulated as:
\begin{align}
\max_{x_{\text{adv}}} \quad 
& \sum_{i=1}^{N} \left\| f^v_i(x_{\text{adv}}) - f^v_i(x) \right\|_2^2 \notag \\
\text{s.t.} \quad 
& \left\| x_{\text{adv}} - x \right\|_{\infty} \leq \epsilon
\label{eq:ssa-cwa}
\end{align}
where $\epsilon$ denotes the maximum allowed perturbation magnitude. This objective aims to generate adversarial examples $x_{adv}$that deviate significantly in feature space across multiple surrogate encoders while remaining visually similar to the original input $x$ under an $L_\infty$ norm constraint.

\noindent
\textbf{Implementation Details. }
We implemented sentence-level inference-time search using the LPS strategy, where the period symbol (“.”) was employed as the segmentation delimiter. At each decoding step, four candidate continuations were generated, with the maximum number of decoding steps set to 10. All experiments were conducted on a single NVIDIA A100 GPU.

\subsection{Evaluation Results}
\noindent
\textbf{Results on General Hallucination Benchmarks.} 
We conducted experiments on three state-of-the-art Multimodal Large Language Models (MLLMs), namely Qwen 2.5 VL, Llama 3.2 Vision , and Phi 3.5 Vision. To evaluate the effectiveness of our proposed method in addressing hallucination issues, we performed experiments on two widely adopted benchmarks: POPE and CHAIR. The results are summarized in Table \ref{tab:pope_chair}, where the best scores are highlighted in bold. As shown by the experimental results, our proposed LPS method consistently outperforms the baseline across all metrics on both datasets, yielding substantial performance improvements. Compared to the baseline that uses the CLIP model as the process reward model, our approach achieves over a 1\% absolute gain on Qwen 2.5 VL. Moreover, it demonstrates significant advantages on the other two models as well. However, we also observe that for the LLaMA 3.2 Vision model, the performance gains of our method on the popular and random categories of the POPE dataset are relatively marginal. We hypothesize that this may be due to the fact that the proposed local perceptual sensitivity is more impactful when the model is under adversarial attacks. In less challenging scenarios, where visual modality features are less perturbed, the benefits of such capability may be less pronounced.
% \vspace{-2mm}
\begin{table}[htb]
  \centering
  {
    \resizebox{0.48\textwidth}{!}{
      \begin{tabular}{c|c|c|c|c}
        \toprule
        \multirow{2}{*}{Step Count} & \multirow{2}{*}{Method} & \multicolumn{3}{c}{Model} \\
        \cline{3-5}
        ~ & ~ & Qwen 2.5 VL & Llama 3.2 Vision & Phi 3.5 Vision \\
        \hline
        \multirow{2}{*}{100} 
        & CLIP PRM & 47.7 & 78.0 & 61.4 \\
        & LPS      & 49.5 & 80.6 & 65.4 \\
        \hline
        \multirow{2}{*}{500} 
        & CLIP PRM & 48.8 & 78.3 & 61.5 \\
        & LPS      & 49.3 & 78.4 & 64.2 \\
        \bottomrule
      \end{tabular}
    }
    \caption{\label{tab:multitrust}Multitrust results across different vision-language models and decoding strategies.}
  }
\end{table}
% \vspace{-2mm}

\noindent
\textbf{Results on Attacked Hallucination Benchmarks. }
To further evaluate the robustness of our method under visual modality attacks, we follow the Multitrust setting and extend its untargeted attack dataset from an original size of 100 samples to a larger set of 1,000 samples. Additionally, to investigate the impact of varying attack intensities on model performance, we conduct experiments using attack steps of 100 and 500. The results are presented in Table \ref{tab:multitrust}. From the results, we observe that our method consistently outperforms the baseline across all attack configurations. In particular, for Phi 3.5 Vision, our method achieves nearly a 3\% performance improvement under both attack step settings. Notably, we also find that the performance of the CLIP-based PRM baseline remains largely unchanged between attack steps of 100 and 500, indicating that this method is highly vulnerable to untargeted attacks—even mild perturbations can induce irreversible degradation in performance. In contrast, our proposed LPS method demonstrates performance gains under both mild and severe attack settings, clearly highlighting its robustness in adversarial scenarios.

\subsection{Further Analysis}

\noindent
\textbf{Comparsion with Other decoding method.}
% \vspace{2mm}
\begin{figure}[htbp]
  \centering
  \includegraphics[width=0.96\linewidth]{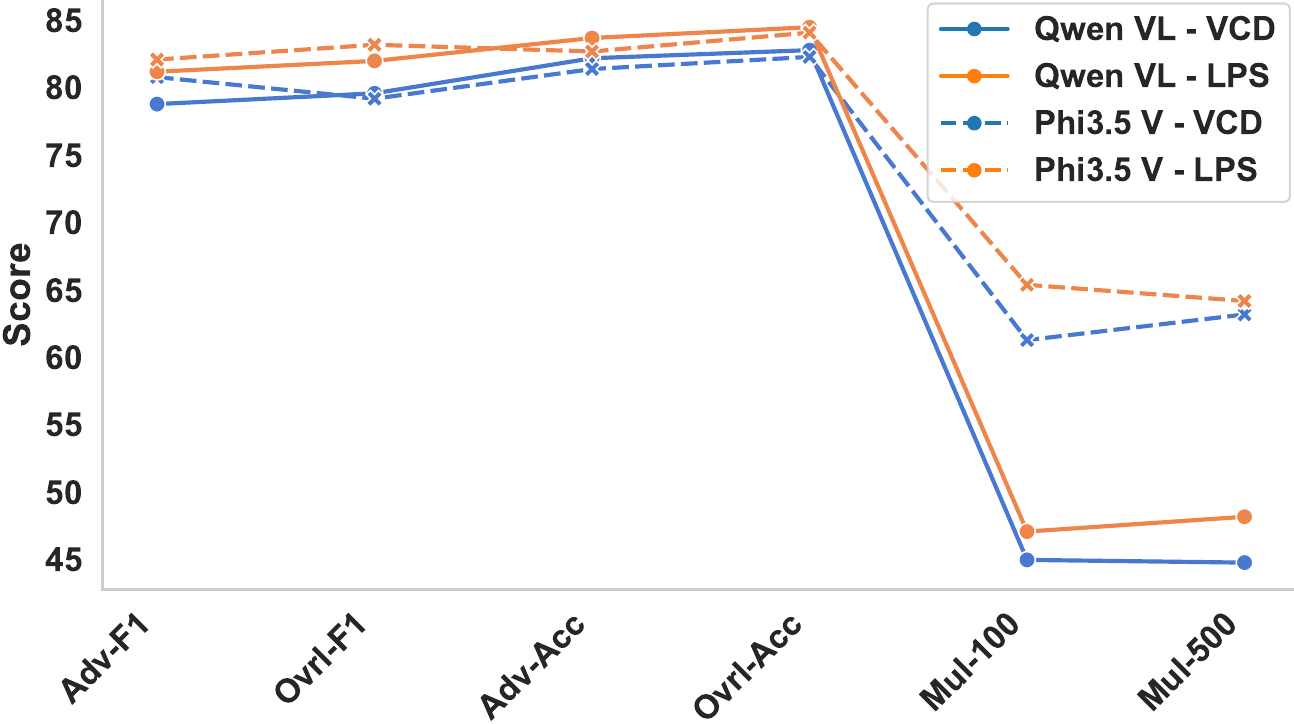}
  \caption{Performance comparison between VCD and LPS on the POPE and Multitrust datasets. "Adv" refers to adversarial performance, while "Ovrl" indicates overall performance. "Mul-100" and "Mul-500" correspond to Multitrust under 100-step and 500-step adversarial attacks, respectively.}
  \label{fig:vcd}
  % \vspace{-2mm}

\end{figure}
% \vspace{-2mm}
To demonstrate the effectiveness of our proposed LPS method, we compare it with the state-of-the-art approach VCD. Experiments are conducted on both the POPE dataset and our 
constructed Multitrust dataset. As shown in Fig. \ref{fig:vcd}, the results indicate that our method consistently outperforms VCD across different models, including Qwen 2.5 VL and Phi 3.5 Vision. Notably, on the Multitrust dataset, our method achieves a significant performance lead. These results strongly suggest that our approach exhibits superior robustness against adversarial attacks compared to VCD.

\begin{figure}[htbp]
  \centering
  \includegraphics[width=1.0\linewidth]{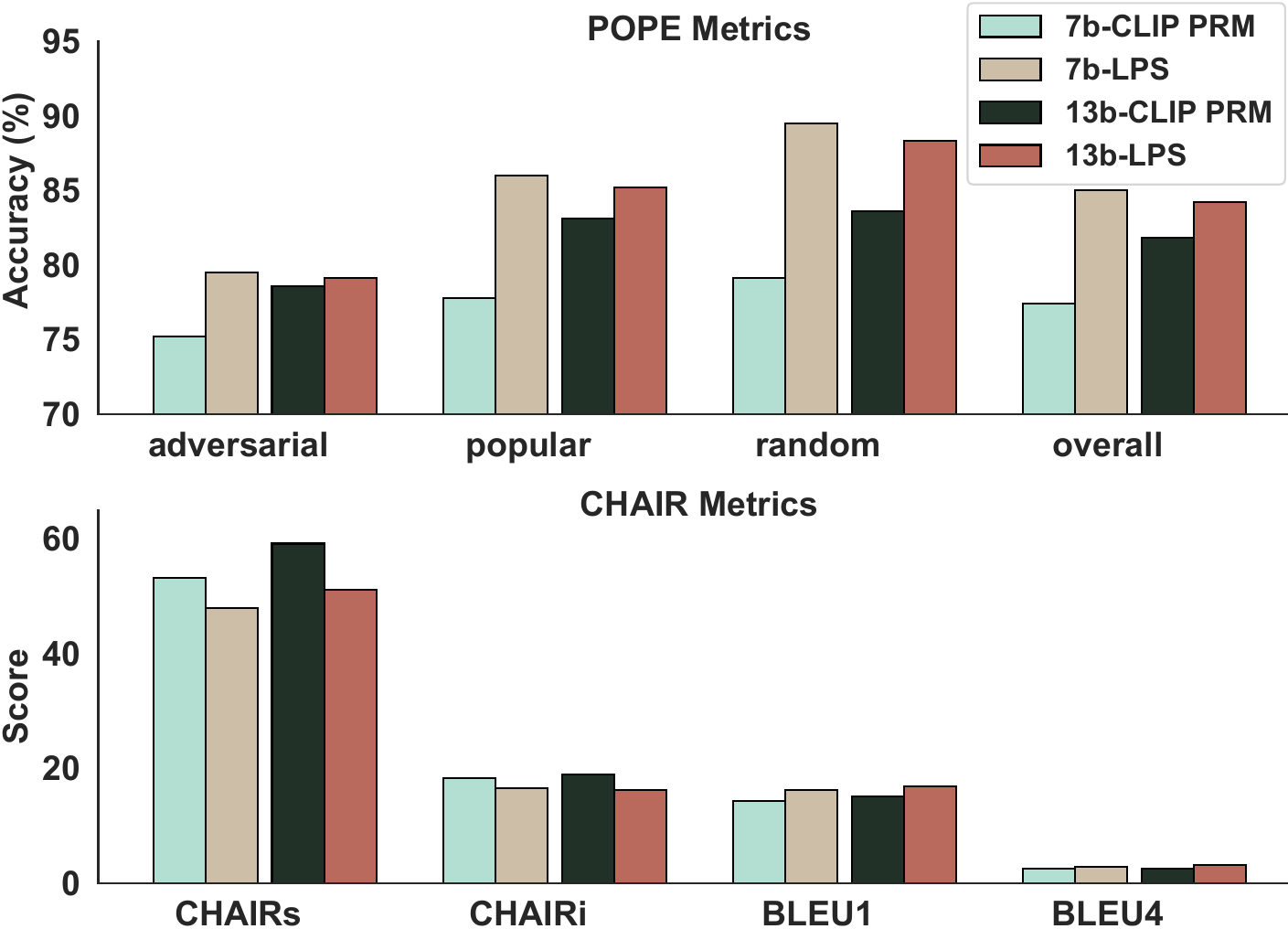}
  \caption{Performance comparison of CLIP-PRM and LPS on LLaVA 1.5 models with different parameter scales (7B vs. 13B) across POPE and CHAIR datasets.}
  \label{fig:llava_pope}
\end{figure}
% \vspace{-2mm}

\begin{figure*}[htb]
  \centering
  \includegraphics[width=0.96\linewidth]{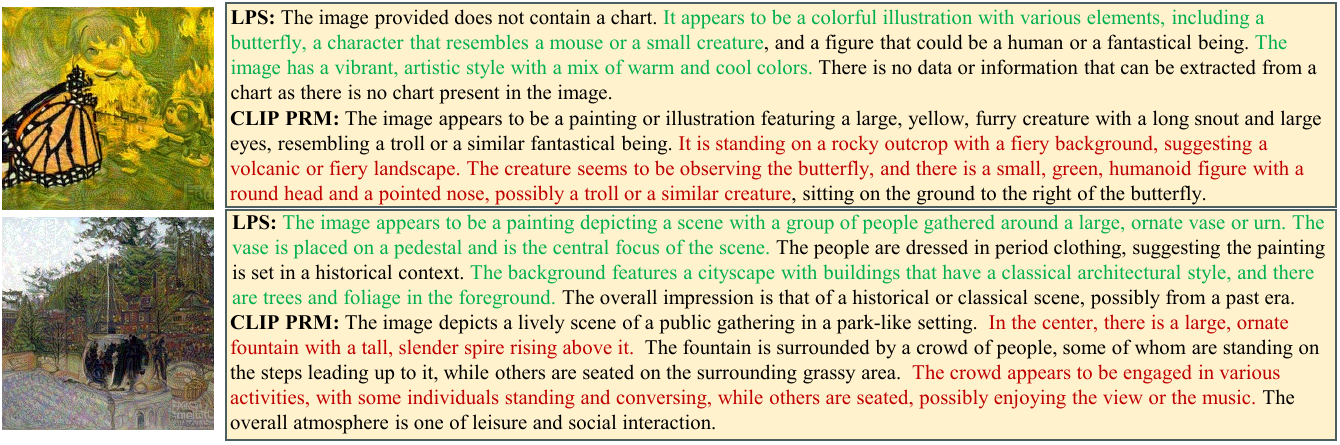}
  \caption{Qualitative comparison of LPS and CLIP-PRM methods when handling adversarially perturbed images. The results show the full response of both methods, highlighting the robustness of LPS against visual perturbations.}
  \label{fig:qa}
\end{figure*}

\noindent
\textbf{Comparison Across Models with Different Parameter Scales. }
To further validate the generality of our proposed LPS method, we conduct a comparative study across models with varying parameter scales within the same architecture family. Specifically, we evaluate LLaVA 1.5-7B and LLaVA 1.5-13B on the POPE and CHAIR datasets, applying both the baseline CLIP-PRM and our LPS approach. As shown in Fig. \ref{fig:llava_pope}, LPS consistently outperforms CLIP-PRM on both datasets across the two model sizes. Notably, on the POPE dataset, LPS achieves more than a 2\% improvement over CLIP-PRM across all evaluation metrics, regardless of whether the underlying model is 7B or 13B. This consistent improvement clearly demonstrates that LPS is effective not only across different model architectures but also remarkably robust to changes in model capacity, further highlighting its scalability and broad general applicability.

\begin{table}[htb]
  % \vspace{-2mm}
  \centering
  {
    \resizebox{0.48\textwidth}{!}{
      \begin{tabular}{c|ccccc}
        \toprule
        \multirow{2}{*}{Model} & \multicolumn{5}{c}{Candidate Number} \\
        \cline{2-6}
        ~ & 1 & 2 & 4 & 6 & 8 \\
        \hline
        Qwen 2.5 VL     & 46.8 & 48.1 & 47.7 & 49.5 & 49.3 \\
        \hline
        LLaMA 3.2 Vision  & 76.1 & 77.3 & 78.0 & 80.6 & 80.3 \\
        \hline
        Phi 3.5 Vision  & 62.2 & 63.7 & 62.9 & 65.4 & 63.8 \\
        \bottomrule
      \end{tabular}
    }
    \caption{\label{tab:candidate_ablation}Ablation study on candidate number for Qwen 2.5 VL, LLaMA 3.2 Vision, and Phi 3.5 Vision on the Multitrust dataset.}
  }
\end{table}
% \vspace{-2mm}

\noindent
\textbf{Ablation Study on Candidate Number. }
To evaluate the effectiveness of our proposed algorithm, we conducted an ablation study on the number of candidates considered at each step. Specifically, we tested model performance with candidate counts ranging from 1 to 8. As shown in Table \ref{tab:candidate_ablation}, the results indicate that for nearly all models, performance consistently improves as the number of candidates per timestep increases. This trend supports the effectiveness of the proposed Local Perception prior in selecting plausible candidates. Additionally, the performance gains gradually diminish as the candidate count continues to rise. We hypothesize that this is because model performance on the task has an inherent upper bound, and as the model approaches this limit, further improvements become increasingly less pronounced.

\noindent
\textbf{Qualitative Analysis. }
To provide a more intuitive understanding of the effectiveness of our method, we conducted a qualitative analysis, as illustrated in Fig. \ref{fig:qa}. In the figure, we present full responses generated by different methods for the image captioning task under adversarial attacks. Correct descriptions are highlighted in green, while hallucinated or incorrect descriptions are marked in red. From the results, it is evident that our proposed method, LPS, is capable of accurately identifying and describing objects—such as butterflies—even under severe visual distortions. In contrast, the baseline method, CLIP-PRM, struggles to interpret adversarially perturbed images, often resorting to superficial cues such as color or texture, and fails to make accurate content-level judgments.

\section{Conclusion}
We propose a novel decoding method called Local Perception Search (LPS), which leverages Local Perception priors to guide the decoding process in Multimodal Large Language Models (MLLMs), aiming to mitigate hallucinations under various conditions—especially when the visual modality is subject to adversarial attacks. We first provide empirical evidence demonstrating the existence of local perception capabilities within MLLMs. Building on this observation, we enhance and utilize this inherent ability by allowing the model to generate its own priors to guide decoding, thereby avoiding reliance on external models or additional components. We evaluate the effectiveness of our approach on widely adopted hallucination benchmarks, POPE and CHAIR, under standard settings. Furthermore, we adopt and extend the untargeted attack dataset from Multitrust to assess performance under visually perturbed inputs. Extensive experiments and comparisons with baselines such as CLIP-PRM and other decoding strategies demonstrate the superior performance of our method.

\section*{Limitations}
The limitations of this study primarily lie in two aspects. First, although our approach eliminates the need for additional components such as grounding detectors, it still relies on the CLIP model to provide vision-language similarity priors. Second, obtaining the Local Perception prior requires an additional inference step, which incurs extra computational and time overhead. These limitations also highlight potential directions for future research.

% \section*{Acknowledgments}

% This document has been adapted
% by Steven Bethard, Ryan Cotterell and Rui Yan
% from the instructions for earlier ACL and NAACL proceedings, including those for
% ACL 2019 by Douwe Kiela and Ivan Vuli\'{c},
% NAACL 2019 by Stephanie Lukin and Alla Roskovskaya,
% ACL 2018 by Shay Cohen, Kevin Gimpel, and Wei Lu,
% NAACL 2018 by Margaret Mitchell and Stephanie Lukin,
% Bib\TeX{} suggestions for (NA)ACL 2017/2018 from Jason Eisner,
% ACL 2017 by Dan Gildea and Min-Yen Kan,
% NAACL 2017 by Margaret Mitchell,
% ACL 2012 by Maggie Li and Michael White,
% ACL 2010 by Jing-Shin Chang and Philipp Koehn,
% ACL 2008 by Johanna D. Moore, Simone Teufel, James Allan, and Sadaoki Furui,
% ACL 2005 by Hwee Tou Ng and Kemal Oflazer,
% ACL 2002 by Eugene Charniak and Dekang Lin,
% and earlier ACL and EACL formats written by several people, including
% John Chen, Henry S. Thompson and Donald Walker.
% Additional elements were taken from the formatting instructions of the \emph{International Joint Conference on Artificial Intelligence} and the \emph{Conference on Computer Vision and Pattern Recognition}.
\bibliography{custom}

\appendix

\section{Details on Experiment Settings}
\label{sec:appendix}

\subsection{Metrics on Hallucination Rate}
\paragraph{(1) CHAIR.} \textit{Caption Hallucination Assessment with Image Relevance (CHAIR)}~ \cite{rohrbach2019objecthallucinationimagecaptioning} proposes a widely adopted metric for hallucination evaluation. This metric assesses hallucination by computing the proportion of referenced objects that do not actually exist in the image, relative to the total number of objects mentioned in the model's output. It includes two variants: \textbf{CHAIR-$\text{S}$}, which evaluates hallucination at the sentence level, and \textbf{CHAIR-$\text{I}$}, which operates at the object-instance level. These two formulations offer complementary perspectives for capturing object hallucination phenomena:

\begin{equation}
\text{CHAIR}_\text{I} = \frac{|\{\text{hallucinated objects}\}|}{|\{\text{all objects}\}|},
\label{eq:chairi}
\end{equation}

\begin{equation}
\text{CHAIR}_\text{S} = \frac{|\{\text{hallucinated responses}\}|}{|\{\text{all responses}\}|},
\label{eq:chairs}
\end{equation}
where hallucinated responses refer to the responses containing at least one hallucinated object.

To assess how closely the model-generated captions align with human-authored ones, we employ the BLEU score~\cite{papineni2002bleu}, a standard metric for evaluating textual similarity. BLEU quantifies the degree of overlap between the model's output and reference captions, thereby reflecting the extent to which the model captures human-like linguistic fidelity and adheres to expected descriptive standards.
\paragraph{(2) POPE.} Consistent with prior studies, our evaluation incorporates the \textit{Polling-based Object Probing Evaluation (POPE)} methodology~\cite{li2023evaluatingobjecthallucinationlarge}. POPE employs an automated segmentation system to identify and outline objects within an image. It then queries the model about the existence of these detected objects, while also introducing randomly selected fictitious objects to assess false positives. The resulting F1 scores provide a comprehensive measure of the model’s ability to accurately perceive and interpret visual content.

\paragraph{(3) Multitrust.} We extend the untargeted attack subset in the Multitrust dataset \cite{zhang2024multitrust} from 100 to 1,000 images to enable a more comprehensive evaluation under visual modality corruption. For each image captioning instance, we examine the model's response to determine whether it mentions any ground truth objects present in the image. A response that explicitly includes at least one ground truth object is considered accurate, while the complete absence of such references is treated as a hallucination.

\subsection{Hyperparameter Settings}
In Equation~\ref{equal:hyper}, we use two hyperparameters, \( a \) and \( b \), to control the balance between the two reward components: $R_{\text{LPS}}^t(i)$ and $R_{\text{CLIP}}^t(i)$. In our experiments, both hyperparameters are set to \( a = 1 \) and \( b = 1 \). In addition, we generate 6 candidate tokens at each decoding step, and set the maximum search step to 10.
\end{document}